\pdfoutput=1
\PassOptionsToPackage{table,dvipsnames}{xcolor}

\documentclass{bmvc2k}

\usepackage{multirow}

\usepackage{microtype}

\usepackage{amssymb}

\usepackage{cleveref}
\crefname{section}{Sec.}{Secs.}
\Crefname{section}{Section}{Sections}
\crefname{table}{Tab.}{Tabs.}
\Crefname{table}{Table}{Tables}
\crefname{figure}{Fig.}{Figs.}
\Crefname{figure}{Figure}{Figures}
\crefname{equation}{Eq.}{Eqs.}

\usepackage{bm}
\usepackage{booktabs}
\renewcommand{\toprule}{\specialrule{1.2pt}{0pt}{2pt}}
\renewcommand{\bottomrule}{\specialrule{1.2pt}{2pt}{8pt}}


\usepackage{tikz}
\usetikzlibrary{arrows.meta,calc,backgrounds,positioning,fit}
\usepackage{pgfplots}
\pgfplotsset{compat=1.18}
\usepgfplotslibrary{groupplots}
\usepackage{graphicx}
\usepackage{stmaryrd}

\definecolor{blueL}{RGB}{218,232,252}   %
\definecolor{blueD}{RGB}{108,142,191}   %
\definecolor{yelL}{RGB}{255,242,204}    %
\definecolor{yelD}{RGB}{241,182,86}     %

\definecolor{boxPair}{RGB}{62,111,193}   %
\definecolor{boxSolo}{RGB}{125,132,142}  %

\definecolor{cOurs}{HTML}{1f77b4}        %
\definecolor{cNaive}{HTML}{d62728}       %
\definecolor{cNone}{HTML}{555555}        %

\crefname{appendix}{Appendix}{Appendices}
\Crefname{appendix}{Appendix}{Appendices}

\title{MOTIP2: Spatial Priors for End-to-End Multi-Object Tracking}

\addauthor{Beno\^{i}t Roussel}{benoit.roussel@ec-lyon.fr}{1,2}
\addauthor{Damien Bouet}{damien.bouet@idemia.com}{2}
\addauthor{Liming Chen}{liming.chen@ec-lyon.fr}{1}
\addauthor{Pierre Perrault}{pierre.perrault@idemia.com}{2}

\addinstitution{
 \'Ecole Centrale de Lyon\\
 LIRIS, UMR CNRS 5205\\
 Lyon, France
}
\addinstitution{
 IDEMIA\\
 Courbevoie, France
}

\runninghead{Roussel \bmvaEtAl}{MOTIP2}

\begin{document}

\maketitle
\makeatletter
{\long\def\@makefntext#1{\noindent#1}%
\footnotetext{Accepted at the British Machine Vision Conference (BMVC) 2026.}}
\makeatother

\begin{abstract}

End-to-end multi-object trackers have narrowed the gap with classical tracking-by-detection on association-difficult benchmarks. Yet they still make spatially implausible errors no classical tracker would, such as assigning one identity to objects on opposite sides of the frame. A model could learn to avoid them, but tracking annotations are scarce, so we encode spatial priors explicitly instead, while keeping inference fully end-to-end with no post-hoc association.

We propose three \emph{spatial priors}, at the data, loss, and representation stages. \emph{Spatial ID Switches} bias trajectory permutations toward spatially overlapping objects, reducing the mismatch between training and inference confusions. \emph{Spatial ID Loss} scales each identity's penalty by its box distance, so a distant switch costs more than a nearby one. \looseness=-1 \emph{Spatial Anchor} gives each track token its frame position, an explicit spatial cue for attention.

We instantiate the three priors in MOTIP2, a tracker adapted from MOTIP and built on the real-time DEIM detection transformer. Trained without extra data, its main model, MOTIP2-L, sets a new state of the art: $\mathbf{73.4}$ HOTA on DanceTrack, $\mathbf{76.0}$ on SportsMOT, and $\mathbf{71.1}$ IDF1 on PersonPath22. MOTIP2 is a family of models spanning the speed-accuracy trade-off: a lighter model, MOTIP2-S, matches the original MOTIP at over $\mathbf{3\times}$ the speed, and MOTIP2-X reaches $\mathbf{74.8}$ HOTA on DanceTrack.
\end{abstract}

\section{Introduction}
\label{sec:intro}

Multi-object tracking supports autonomous driving~\cite{BDD100K}, sports analytics~\cite{SportsMOT}, and video surveillance~\cite{PersonPath22}, all of which need stable identities through occlusion and crowding. End-to-end multi-object trackers~\cite{MOTR, TrackFormer, MOTIP} extend the DETR~\cite{DETR} query paradigm to tracking: detection without NMS, and learned association replacing the hand-crafted IoU matching that breaks down under heavy occlusion. They match or surpass classical tracking-by-detection~\cite{SORT, ByteTrack, OC-SORT, BoT-SORT} on association-difficult benchmarks such as DanceTrack~\cite{DanceTrack}.

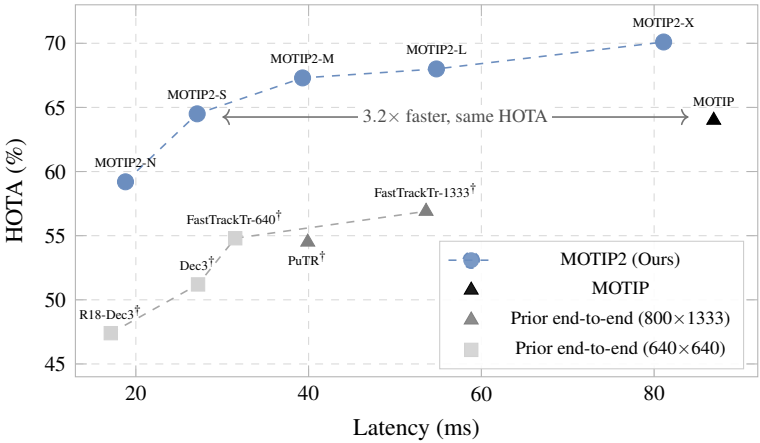
\begin{figure}[t]
    \centering
    \begin{tikzpicture}
        \begin{axis}[
            width=0.82\linewidth,
            height=6.5cm,
            xlabel={Latency (ms)},
            ylabel={HOTA (\%)},
            xlabel style={font=\small},
            ylabel style={font=\small},
            xmin=13, xmax=92,
            ymin=44, ymax=73,
            xtick={20,40,60,80},
            ytick={45,50,55,60,65,70},
            tick label style={font=\footnotesize},
            legend style={
                at={(0.98,0.02)},
                anchor=south east,
                font=\scriptsize,
                draw=gray!40,
                fill=white,
                fill opacity=0.9,
                text opacity=1,
                row sep=1pt,
                column sep=4pt,
            },
            grid=major,
            grid style={dashed, gray!30},
            axis line style={gray!60},
            tick style={gray!60},
            clip=false,
        ]

            \addplot[
                gray!70,
                semithick,
                dashed,
                mark=none,
                forget plot,
            ] coordinates {
                (17.1,47.4)
                (27.2,51.2)
                (31.5,54.8)
                (53.6,56.9)
            };

            \addplot[
                blueD,
                semithick,
                dashed,
                mark=none,
                forget plot,
            ] coordinates {
                (18.8,59.2)
                (27.1,64.5)
                (39.3,67.3)
                (54.8,68.0)
                (81.1,70.1)
            };

            \addplot[
                only marks,
                mark=triangle*,
                mark size=3pt,
                draw=gray,
                fill=gray,
                forget plot,
            ] coordinates {
                (39.9,54.5)
            };

            \addplot[
                only marks,
                mark=triangle*,
                mark size=3pt,
                draw=gray,
                fill=gray,
                forget plot,
            ] coordinates {
                (53.6,56.9)
            };

            \addplot[
                only marks,
                mark=triangle*,
                mark size=3pt,
                draw=black,
                fill=black,
                forget plot,
            ] coordinates {
                (86.9,64.0)
            };

            \addplot[
                only marks,
                mark=square*,
                mark size=2.5pt,
                draw=gray!35,
                fill=gray!35,
                forget plot,
            ] coordinates {
                (17.1,47.4)
                (27.2,51.2)
                (31.5,54.8)
            };

            \addplot[
                only marks,
                mark=*,
                mark size=3pt,
                draw=blueD,
                fill=blueD,
                forget plot,
            ] coordinates {
                (18.8,59.2)
                (27.1,64.5)
                (39.3,67.3)
                (54.8,68.0)
                (81.1,70.1)
            };

            \node[font=\tiny, black, anchor=south] at (axis cs:17.1,47.8) {R18-Dec3$^\dagger$};
            \node[font=\tiny, black, anchor=south] at (axis cs:27.2,51.6) {Dec3$^\dagger$};
            \node[font=\tiny, black, anchor=south] at (axis cs:31.5,55.2) {FastTrackTr-640$^\dagger$};
            \node[font=\tiny, black, anchor=north] at (axis cs:39.9,54.5) {PuTR$^\dagger$};
            \node[font=\tiny, black, anchor=south] at (axis cs:53.6,57.3) {FastTrackTr-1333$^\dagger$};
            \node[font=\tiny, black, anchor=south] at (axis cs:86.9,64.4) {MOTIP};

            \node[font=\tiny, black, anchor=south] at (axis cs:18.8,59.6) {MOTIP2-N};
            \node[font=\tiny, black, anchor=south] at (axis cs:27.1,64.9) {MOTIP2-S};
            \node[font=\tiny, black, anchor=south] at (axis cs:39.3,67.7) {MOTIP2-M};
            \node[font=\tiny, black, anchor=south] at (axis cs:54.8,68.4) {MOTIP2-L};
            \node[font=\tiny, black, anchor=south] at (axis cs:81.1,70.5) {MOTIP2-X};

            \draw[<->, semithick, black!55]
                (axis cs:30.0,64.25) -- (axis cs:84.0,64.25);
            \node[font=\scriptsize, black!70, fill=white, inner sep=1.5pt]
                at (axis cs:57,64.25) {$3.2\times$ faster, same HOTA};

            \addlegendimage{blueD, semithick, dashed, mark=*, mark size=3pt, mark options={draw=blueD, fill=blueD}}
            \addlegendentry{MOTIP2 (Ours)}

            \addlegendimage{only marks, mark=triangle*, mark size=3pt, draw=black, fill=black}
            \addlegendentry{MOTIP}

            \addlegendimage{only marks, mark=triangle*, mark size=3pt, draw=gray, fill=gray}
            \addlegendentry{Prior end-to-end (800$\times$1333)}

            \addlegendimage{only marks, mark=square*, mark size=2.5pt, draw=gray!35, fill=gray!35}
            \addlegendentry{Prior end-to-end (640$\times$640)}

        \end{axis}
    \end{tikzpicture}
    \caption{\textbf{Accuracy vs.\ latency.} HOTA on DanceTrack-val vs.\ per-frame PyTorch FP32 latency (ms) on a single V100. At matched accuracy, MOTIP2-S is $3.2\times$ faster than MOTIP. Marker shape denotes input resolution: square 640$\times$640, gray triangle 800$\times$1333; MOTIP (black) and MOTIP2 (circles) run at 800$\times$1440. $^\dagger$~Estimated from RTX 4090 measurements~\cite{FastTrackTr} scaled by MOTIP's V100/4090 ratio.}
    \label{fig:hota_fps}
\end{figure}

Two recurring failure types, however, share one cause: association without spatial grounding. In the first, two distant boxes in consecutive frames, with no occlusion between them, are assigned the same identity (see \cref{sec:suppl_longswitch}), a failure rare in classical trackers, whose motion gating makes such jumps expensive. In the second, the tracker swaps identities when objects cross or occlude. Neither is constrained by geometry: the cross-entropy penalizes an implausible cross-frame jump no more than an adjacent mistake, and training barely exposes the model to the nearby confusions where association is genuinely ambiguous. The association step needs an explicit spatial prior.

Tracking-by-detection encodes such constraints for free, as hand-crafted heuristics that need no training data: Kalman motion, IoU cost, and overlap-gated re-identification each forbid an association the moment it becomes spatially implausible~\cite{DeepSORT, BoT-SORT}. End-to-end trackers discard these components and leave the attention mechanism to learn the same constraints from data. But annotated tracking data is scarce, so these constraints are learned imperfectly, and the tracker makes the implausible associations they would have prevented.

We propose three \emph{explicit spatial priors} to tackle both failures, one each at the data, loss, and representation stage, all encoding one bias: spatial proximity should govern association. At the data stage, \emph{Spatial ID Switches} target the dominant occlusion confusions. MOTIP exposes its decoder to such errors by shuffling historical identities at random, but a random shuffle mostly fabricates long-range swaps that rarely occur at inference, while barely sampling the nearby occlusions that cause most errors. We instead bias permutations toward spatially overlapping objects, so augmented histories mirror the confusions the tracker actually encounters. This trains the decoder to look further back in the tracklet to disambiguate when nearby objects are present, rather than blindly trusting the most recent assignment (\cref{fig:idsw_qual}).

At the loss stage, the \emph{Spatial ID Loss} targets the distant jump by restoring the geometric cost the cross-entropy leaves out. At the representation stage, the \emph{Spatial Anchor} ties each track token to its frame position, a dedicated channel through which geometry enters association, so the model discriminates between candidates by where they are rather than only penalizing incoherent associations after the fact. Unlike the heuristic stack, these priors add no hand-crafted matching at test time, leaving inference free of any post-hoc association module.

We instantiate the three priors in \emph{MOTIP2}, an end-to-end tracker built on MOTIP~\cite{MOTIP} with a real-time DEIM~\cite{DEIM} detection backbone and an adapted ID decoder, and evaluate it on DanceTrack~\cite{DanceTrack}, SportsMOT~\cite{SportsMOT}, and PersonPath22~\cite{PersonPath22}. The gains separate cleanly: the DEIM backbone drives detection and speed, while under a fixed backbone the priors alone add up to $+2.1$ HOTA and $+3.5$ AssA on DanceTrack test (\cref{tab:sota_dancetrack}). Transformer trackers have long been latency-bound; MOTIP2 runs at real-time rates, over $\mathbf{3\times}$ faster than MOTIP at matched accuracy (\cref{fig:hota_fps}).

\paragraph{Contributions.}

\begin{itemize}\setlength{\itemsep}{2pt}
    \item[$(i)$] \textbf{Spatial ID Switches}: a data-stage augmentation that replaces uniform tracklet shuffling with overlap-weighted permutations.

    \item[$(ii)$] \textbf{Spatial ID Loss}: a loss-stage term that scales each candidate identity's penalty by its box distance to the detection.

    \item[$(iii)$] \textbf{Spatial Anchor}: a representation-stage change that reinjects the detector's positional encoding into the ID decoder, reusing a spatial cue it already computes.

    \item[$(iv)$] \textbf{MOTIP2}: a real-time, DEIM-based end-to-end tracker, adapted from MOTIP, that validates the three priors with consistent gains across three tracking benchmarks.
\end{itemize}

\section{Related Work}
\label{sec:related_work}

\paragraph{Tracking-by-detection.}
Tracking-by-detection pipelines~\cite{SORT, DeepSORT, ByteTrack, OC-SORT, BoT-SORT} pair a frame-level detector~\cite{YoloX} with a small, hand-designed association stack: a Kalman filter~\cite{KalmanFilter} propagates each track forward, IoU cost penalizes implausible jumps, and Hungarian matching~\cite{Hungarian} resolves the assignment. When appearance is added via a re-identification network~\cite{DeepSORT, FairMOT, JDE, BoT-SORT, Hybrid-SORT}, its score is gated by motion or overlap rather than consulted in isolation. Every component carries an explicit spatial prior, and the association module itself has essentially no learned parameters; this kept the pipeline competitive while annotated tracking data stayed scarce. These pipelines also commonly delay committing tentative tracks until they are observed across several consecutive frames~\cite{DeepSORT}, a tracklet-confirmation mechanism that prior query-based trackers have not adopted. The stack fails when crowding makes these geometric priors ambiguous: motion gating and IoU cost both lose their discriminative power as the IoU between an object's consecutive observations collapses toward the inter-object IoU with its neighbors. This limitation motivated learned association.

\paragraph{Detection Transformers.}
DETR~\cite{DETR} reformulates detection as set prediction on the transformer~\cite{Transformer}, dispensing with the NMS step that suppresses valid detections when targets overlap. A line of follow-up work progressively re-grounds the decoder in geometry: Deformable DETR~\cite{DeformableDETR} attaches reference points to each query, DAB-DETR~\cite{DAB-DETR} represents queries explicitly as 4D anchor boxes, and DN-DETR~\cite{DN-DETR} and DINO~\cite{DINO} add box denoising objectives that anchor queries against ground-truth geometry during training. Real-time variants~\cite{RT-DETR, D-FINE, DEIM} bring this family to tracking-by-detection latencies while retaining the anchored-query structure; we build on DEIM~\cite{DEIM} for its accuracy at real-time latency.

\paragraph{Tracking-by-query.}
Transformer-based trackers~\cite{TransTrack, TrackFormer, MOTR} inherit DETR's set-prediction formulation and add temporal association by propagating object queries across frames. MOTR~\cite{MOTR} formalizes this as a fully end-to-end framework, but the decoder it shares between detection and association creates a supervision conflict that MOTRv2~\cite{MOTRv2}, MOTRv3~\cite{MOTRv3}, and CO-MOT~\cite{CO-MOT} address through query-specific supervision or external proposals. MOTIP~\cite{MOTIP} resolves the conflict architecturally: a dedicated ID decoder predicts identity labels from the detector's outputs, decoupling the two tasks and enabling scalable training, with trajectory-level identity permutations as the main association augmentation. The ID decoder, however, consumes only the detector's content embeddings: the explicit per-query anchor that DAB-DETR and its successors introduced inside the detector is not carried through to association, leaving the module to recover spatial coherence from content alone. FDTA~\cite{FDTA}, also building on MOTIP, instead learns more discriminative object embeddings for association and is our primary baseline (\cref{sec:sota}). Transformer trackers also remain substantially slower than tracking-by-detection, a gap that FastTrackTr~\cite{FastTrackTr} begins to close. Concurrent work raises end-to-end association accuracy through segmentation-driven tracking built on foundation models~\cite{SAM2MOT}, a direction complementary to the spatial priors we introduce. Our work targets both gaps, spatial grounding and speed: we restore the missing spatial prior at three stages of the association pipeline (data, loss, representation) on a real-time DEIM backbone.

\section{Spatial Priors for ID Association}
\label{sec:method}

\begin{figure*}[t]
\centering
\includegraphics[width=0.9\textwidth]{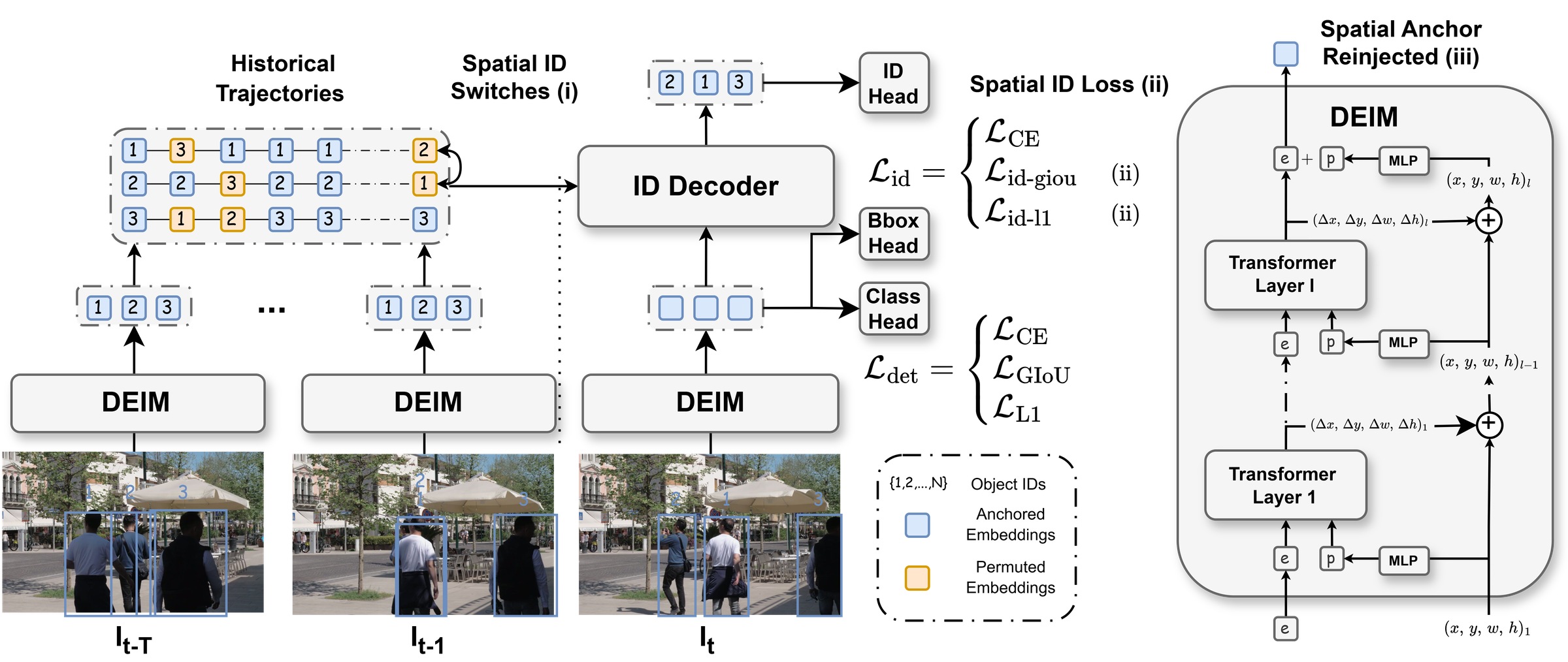}
\caption{
\textbf{Overview of MOTIP2.}
DEIM produces detection embeddings independently per frame. Three explicit spatial priors then inject geometric structure at distinct stages: \textbf{(i) Spatial ID Switches} perturb the training-time historical context by GIoU-biased identity permutations that mirror the near-track confusions seen at inference; \textbf{(ii) Spatial ID Loss} supervises the ID decoder with a geometric penalty; \textbf{(iii) Spatial Anchor} carries each detection's positional encoding into the ID decoder, so every token holds an explicit cue to its image position.}
\label{fig:overview}
\end{figure*}

We first present MOTIP2, our end-to-end tracker adapted from MOTIP~\cite{MOTIP}, together with modifications to its detector and ID decoder (\cref{sec:overview}, \cref{fig:overview}). We then instantiate the three spatial priors announced in \cref{sec:intro}, each acting at a different stage of the tracker: \emph{Spatial ID Switches} at the data stage (\cref{sec:id_switches}), the \emph{Spatial ID Loss} at the loss stage (\cref{sec:spatial_loss}), and \emph{Spatial Anchor} at the representation stage (\cref{sec:anchor}). 

\subsection{Overview: MOTIP2}
\label{sec:overview}

MOTIP2 follows MOTIP~\cite{MOTIP}, which we briefly recap; readers familiar with it can skip to the changes. MOTIP decouples detection from identity prediction within a single end-to-end tracker. A detection transformer processes each frame independently, producing detection embeddings $\mathbf{e}_i$ with bounding boxes $\hat{b}_i$. During training, bipartite matching assigns each prediction to a ground-truth object with identity label $y_i$, as in DETR; a dedicated ID decoder then cross-attends over a historical context of the last $T$ frames and predicts, for each detection, a distribution over an \emph{ID dictionary} of $K{+}1$ learnable identity embeddings ($K$ active identities plus a newborn class). The model is trained end-to-end on short clips sampled at random frame rates, jointly optimizing the detector and the ID decoder. At inference the video is processed online: each frame's assigned embeddings are appended to the context, maintained as a FIFO buffer of size $T$, and a greedy protocol turns the predicted distributions into unique identity assignments.

Throughout, \emph{end-to-end} means no post-hoc association module rather than no tracking logic at inference: the greedy readout and the tracklet confirmation below manage track lifecycles, with no matching stage between separately produced detections and tracks. Beyond the three spatial priors that motivate this work (\cref{sec:id_switches,sec:spatial_loss,sec:anchor}), our changes are confined to the detector, the ID decoder, and the training objective, described next; everything else above is inherited unchanged.

\paragraph{Detection backbone.}
We replace MOTIP's Deformable DETR detector~\cite{DeformableDETR} with the real-time detector DEIM~\cite{DEIM}. Profiling shows that over 80\% of MOTIP's inference cost comes from the detector (\cref{tab:latency_trt}), making it the primary target for acceleration. Selecting different DEIM backbone and decoder configurations instantiates a family of models (N/S/M/L/X): even the high-capacity X variant matches MOTIP's speed while being more accurate, and lighter variants cut latency further.

\paragraph{ID decoder.}
We keep MOTIP's ID decoder with two structural changes. First, we fuse detection features and identity embeddings by element-wise addition rather than concatenation, halving the hidden dimension to match the DETR decoder; we double the feed-forward network (FFN) ratio to preserve capacity. Second, we replace MOTIP's learned discrete temporal attention bias with rotary position embeddings (RoPE)~\cite{RoPE} applied inside the query--key product, so attention between two tokens varies with their frame offset. RoPE enables context-window extension at inference beyond the training length~\cite{PE_Interpolation}; because training samples clips at variable frame rates while inference runs at a fixed stride, we rescale the RoPE timescale at inference by a factor $\alpha$ to match the average training timescale. Finally, at inference we add a lightweight tracklet-confirmation step absent from MOTIP: a track is confirmed only after $N_{\text{init}}{=}2$ consecutive consistent assignments, and unconfirmed tracks are discarded with their dictionary slots freed.

\subsection{Spatial ID Switches}
\label{sec:id_switches}

Because bipartite matching assigns each token to its ground-truth identity, the training-time historical context is, by construction, perfectly consistent. At inference, however, noisy detections and occlusions cause identity ambiguities. To bridge this gap, MOTIP~\cite{MOTIP} perturbs the training context with an ID-switch augmentation that selects a random subset of objects in each frame and permutes their identities among themselves. The ID decoder thus learns to recover from confusions rather than overfitting to a perfect history.

Genuine confusions, however, are local: they arise between nearby objects under occlusion or close passage. Objects on opposite sides of the frame are never genuinely ambiguous, yet a uniform permutation, blind to geometry, fabricates mostly these long-range pairings while under-sampling the nearby confusions that cause most errors.

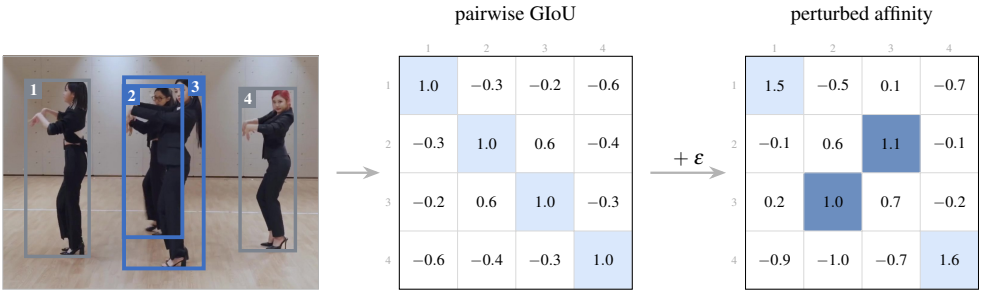
\begin{figure}[t]
\centering
\resizebox{\columnwidth}{!}{%
\begin{tikzpicture}[font=\sffamily, >=Stealth]

\def\cw{0.9}

\def\idx{%
  \foreach \k [count=\n from 0] in {1,2,3,4} {%
    \node[font=\tiny, gray!65, anchor=south, inner sep=1pt]
      at ({\n*\cw+\cw/2},{4*\cw+0.02}) {\k};
    \node[font=\tiny, gray!65, anchor=east, inner sep=2pt]
      at (-0.05,{(4-\n)*\cw-\cw/2}) {\k};
  }%
}

\begin{scope}
  \node[anchor=south west, inner sep=0] at (0,0) {\includegraphics[width=4.877cm]{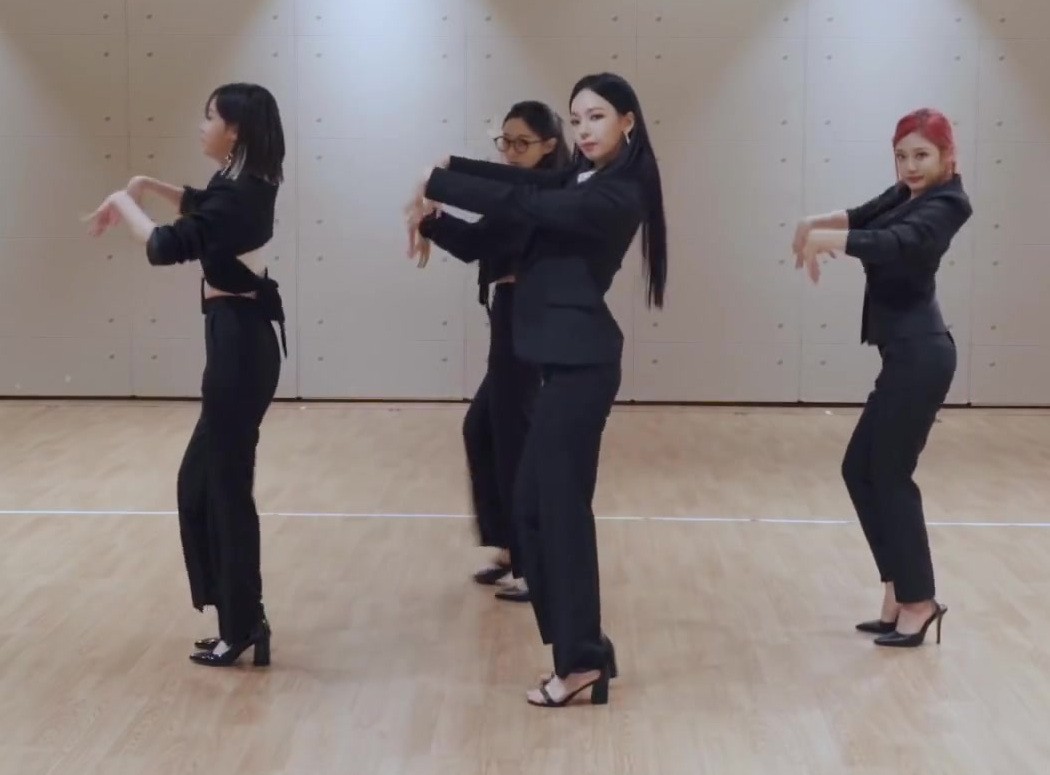}};
  \draw[white!70!black, line width=0.4pt] (0,0) rectangle (4.877,3.600);
  \draw[boxSolo, line width=1.5pt] (0.354,0.507) rectangle (1.334,3.229);
  \node[fill=boxSolo, text=white, font=\scriptsize\bfseries, inner sep=1.6pt, anchor=north west] at (0.354,3.229) {1};
  \draw[boxSolo, line width=1.5pt] (3.666,0.595) rectangle (4.521,3.103);
  \node[fill=boxSolo, text=white, font=\scriptsize\bfseries, inner sep=1.6pt, anchor=north west] at (3.666,3.103) {4};
  \draw[boxPair, line width=1.7pt] (1.868,0.804) rectangle (2.774,3.140);
  \node[fill=boxPair, text=white, font=\scriptsize\bfseries, inner sep=1.6pt, anchor=north west] at (1.868,3.140) {2};
  \draw[boxPair, line width=1.7pt] (1.873,0.321) rectangle (3.104,3.275);
  \node[fill=boxPair, text=white, font=\scriptsize\bfseries, inner sep=1.6pt, anchor=north east] at (3.104,3.275) {3};
\end{scope}

\draw[->, gray!60, line width=1pt] (5.157,{2*\cw}) -- (5.827,{2*\cw});

\begin{scope}[shift={(6.127,0)}]
  \node[anchor=south, font=\small, inner sep=2pt] at ({2*\cw},{4*\cw+0.42}) {pairwise GIoU};
  \fill[blueL] ({0*\cw},{3*\cw}) rectangle ({1*\cw},{4*\cw});
  \fill[blueL] ({1*\cw},{2*\cw}) rectangle ({2*\cw},{3*\cw});
  \fill[blueL] ({2*\cw},{1*\cw}) rectangle ({3*\cw},{2*\cw});
  \fill[blueL] ({3*\cw},{0*\cw}) rectangle ({4*\cw},{1*\cw});
  \draw[gray!30] (0,0) grid[step=\cw] ({4*\cw},{4*\cw});
  \draw[black] (0,0) rectangle ({4*\cw},{4*\cw});
  \idx
  \foreach \i/\j/\v in {
    0/0/1.0, 0/1/{-0.3}, 0/2/{-0.2}, 0/3/{-0.6},
    1/0/{-0.3}, 1/1/1.0, 1/2/0.6, 1/3/{-0.4},
    2/0/{-0.2}, 2/1/0.6, 2/2/1.0, 2/3/{-0.3},
    3/0/{-0.6}, 3/1/{-0.4}, 3/2/{-0.3}, 3/3/1.0}
    \node[font=\scriptsize] at ({\j*\cw+\cw/2},{(3-\i)*\cw+\cw/2}) {$\v$};
\end{scope}

\draw[->, gray!60, line width=1pt] (10.007,{2*\cw}) -- (11.177,{2*\cw});

\node[anchor=south, font=\small, inner sep=3pt] at (10.602,{2*\cw}) {$+\;\varepsilon$};

\begin{scope}[shift={(11.477,0)}]
  \node[anchor=south, font=\small, inner sep=2pt] at ({2*\cw},{4*\cw+0.42}) {perturbed affinity};
  \fill[blueL] ({0*\cw},{3*\cw}) rectangle ({1*\cw},{4*\cw});
  \fill[blueL] ({3*\cw},{0*\cw}) rectangle ({4*\cw},{1*\cw});
  \fill[blueD] ({2*\cw},{2*\cw}) rectangle ({3*\cw},{3*\cw});
  \fill[blueD] ({1*\cw},{1*\cw}) rectangle ({2*\cw},{2*\cw});
  \draw[gray!30] (0,0) grid[step=\cw] ({4*\cw},{4*\cw});
  \draw[black] (0,0) rectangle ({4*\cw},{4*\cw});
  \idx
  \foreach \i/\j/\v in {
    0/0/1.5, 0/1/{-0.5}, 0/2/0.1, 0/3/{-0.7},
    1/0/{-0.1}, 1/1/0.6, 1/2/1.1, 1/3/{-0.1},
    2/0/0.2, 2/1/1.0, 2/2/0.7, 2/3/{-0.2},
    3/0/{-0.9}, 3/1/{-1.0}, 3/2/{-0.7}, 3/3/1.6}
    \node[font=\scriptsize] at ({\j*\cw+\cw/2},{(3-\i)*\cw+\cw/2}) {$\v$};
\end{scope}

\end{tikzpicture}%
}%
\caption{\textbf{Spatial ID Switches.}
A DanceTrack~\cite{DanceTrack} frame with four annotated objects, of which only $2$ and $3$ overlap, and the identity permutation the augmentation draws for this frame. We perturb the pairwise GIoU affinity with i.i.d.\ Gaussian noise $\varepsilon_{ij} \sim \mathcal{N}(0,\,\sigma^2)$ and let the Hungarian algorithm return the assignment maximizing it; cell $(i,j)$ means object $i$ takes identity $j$, light where the identity is kept and dark where it is swapped. Since only spatially close pairs carry positive affinity, the perturbation swaps $2\!\leftrightarrow\!3$ while the well-separated objects keep their identity.}
\label{fig:proximity_matrices}
\end{figure}

We keep this random selection of objects and change only how the selected identities are reassigned, biasing the permutation toward spatially proximate boxes instead of drawing it uniformly. Given the selected boxes in frame $t$, we compute their pairwise GIoU~\cite{gIoU} affinity matrix $G_t$ and perturb it with Gaussian noise:
\[
A_t = G_t + \varepsilon, \quad \varepsilon_{ij} \overset{\text{i.i.d.}}{\sim} \mathcal{N}(0,\,\sigma^2),
\]
where $\sigma > 0$ sets the GIoU scale below which two objects are treated as confusable. The augmented permutation on the random subset is then obtained by solving the linear assignment on $A_t$ via the Hungarian algorithm (\cref{fig:proximity_matrices}).

The augmentation recovers MOTIP's uniform permutation exactly as $\sigma \to \infty$, where noise dominates the affinity. For finite $\sigma$, an object swaps only when a nearby box's perturbed affinity overtakes its self-affinity ($G_{ii}{=}1$), so $\sigma$ is a confusion radius in GIoU space: objects with a close, high-GIoU neighbor face a real swap probability while well-separated objects stay virtually untouched. The augmentation thus adapts to scene geometry on its own, and because locality is fixed by geometry rather than a tuned swap rate, we use a single $\sigma{=}0.5$ on all benchmarks without per-dataset tuning.

\subsection{Spatial ID Loss}
\label{sec:spatial_loss}

Cross-entropy ($\mathcal{L}_{\text{CE}}$) over the ID softmax is spatially blind: a confident wrong prediction on a track across the frame is penalized identically to a confusion with an adjacent track. The gradient carries no geometric signal. We complement it with two spatial terms (L1 and GIoU) that weight each candidate identity by the distance between the detection and that identity's box:
\begin{equation}
  \mathcal{L}_{\text{id-l1}} = \sum_{c} p_c \, \|\hat{b} - b^{\text{gt}}_c\|_1,
  \qquad
  \mathcal{L}_{\text{id-giou}} = \sum_{c} p_c \, \bigl(1 - \mathrm{GIoU}(\hat{b}, b^{\text{gt}}_c)\bigr),
  \label{eq:id_components}
\end{equation}
where $p_c$ is the softmax probability for identity $c$, $\hat{b}$ is the predicted box of the detection being classified, and $b^{\text{gt}}_c$ is the ground-truth box of the object holding identity $c$, taken from the same bipartite matching that supervises detection (\cref{sec:overview}). The sum runs over the identities present in the current frame plus the newborn class: identities absent from the frame have no box, so the probabilities $p_c$ are renormalized over this reduced set rather than taken from the full-dictionary softmax. The newborn class has no associated object, so we set $b^{\text{gt}}_c$ to the detection's own ground-truth box, reducing the term to its localization error and leaving the cross-entropy to separate newborn from the correct existing identity. Because $\hat{b}$ is detached, these terms shape only the ID logits and leave the detection regression head untouched.

The total ID objective adds these two spatial terms to the cross-entropy, weighted by the same $(\lambda_{\text{CE}}, \lambda_{\text{GIoU}}, \lambda_{\text{L1}})$ as the DEIM detection loss:
\begin{equation}
  \mathcal{L}_{\text{id}} = \lambda_{\text{CE}} \, \mathcal{L}_{\text{CE}} + \lambda_{\text{GIoU}} \, \mathcal{L}_{\text{id-giou}} + \lambda_{\text{L1}} \, \mathcal{L}_{\text{id-l1}}.
  \label{eq:id_loss}
\end{equation}

\subsection{Spatial Anchor}
\label{sec:anchor}

Detection in modern DETR variants~\cite{DAB-DETR,DN-DETR,DINO,RT-DETR,DEIM} is built on explicit spatial grounding: each query is a 4D anchor box, refined layer by layer, and a positional encoding $\mathbf{p}_i$ (a learned embedding of the final-layer box) steers cross-attention toward the relevant image region at every decoder step. At the handoff to the ID decoder, however, only the content embedding $\mathbf{e}_i$ is carried forward; $\mathbf{p}_i$ is dropped, leaving the decoder to perform association without any explicit cue about \emph{where} on the frame each track sits, despite this information being available one stage upstream.

End-to-end training could let $\mathbf{e}_i$ encode location implicitly, but only entangled with appearance and re-learned from scratch, whereas the detector already computes this cue explicitly. We reuse it directly.

We retain $\mathbf{p}_i$ at this handoff: it is added to each detection token as it enters the ID decoder, so every track carries its frame position, taken from its frame of origin, into association. Position then travels with the content embedding through the decoder's attention and feed-forward layers, an explicit spatial signal the model would otherwise have to recover from appearance alone.

\section{Experiments}
\label{sec:experiments}

We first describe the training data and evaluation protocol, then compare with state-of-the-art methods across benchmarks spanning varied tracking regimes (\cref{sec:sota}), analyze the speed-accuracy trade-off across model sizes (\cref{sec:speed}), and finally isolate each contribution through an ablation study on PersonPath22 and DanceTrack (\cref{sec:ablations}).

\subsection{Experimental Setup}

\paragraph{Datasets.} We evaluate on three benchmarks that stress different facets of association: DanceTrack~\cite{DanceTrack} (similar appearance with complex non-linear motion), SportsMOT~\cite{SportsMOT} (fast non-linear motion in team sports), and PersonPath22~\cite{PersonPath22} (large-scale, diverse real-world scenes). We train one model per benchmark on its own training split, additionally co-training the PersonPath22 model with CrowdHuman~\cite{CrowdHuman}. We report MOTIP2-L as our default variant for state-of-the-art comparisons unless otherwise specified, with MOTIP2-S used for ablations (\cref{sec:ablations}).

\paragraph{Metrics.}
We primarily report Higher Order Tracking Accuracy (HOTA)~\cite{HOTA}, which balances detection accuracy (DetA) and association accuracy (AssA); we also report MOTA~\cite{CLEAR} and IDF1~\cite{IDF1}.

\paragraph{Network.}
We initialize DEIM~\cite{DEIM} from its official COCO~\cite{COCO} pre-trained weights. The ID decoder mirrors the DEIM decoder's hidden dimension and number of layers, so both components scale together across model variants.

\paragraph{Training.}
We run all experiments on 4 NVIDIA V100 GPUs. Training follows a two-stage schedule: we first pretrain the DEIM detector alone, then jointly train the full model with AdamW~\cite{AdamW}, gradient clipping, and exponential moving average (EMA). Full optimizer, schedule, clip-sampling, and augmentation settings are detailed in \cref{sec:suppl_training}.

\paragraph{Hyperparameters.}
The Spatial ID Loss uses $\lambda_{\text{CE}}{=}1.0$, $\lambda_{\text{GIoU}}{=}2.0$, and $\lambda_{\text{L1}}{=}5.0$, matching the DEIM detection-loss weights. Spatial ID Switches use a confusion radius $\sigma{=}0.5$, tracklet confirmation requires $N_{\text{init}}{=}2$ consecutive frames, greedy ID assignment requires a minimum identity probability $\lambda_{\text{ID}}{=}0.1$, and the RoPE timescale is rescaled by $\alpha{=}2.0$ at inference. These, together with the inference resolution (shorter side $800$ pixels, longer side capped at $1440$, matching both tracking-by-detection~\cite{ByteTrack,BoT-SORT} and end-to-end~\cite{MOTIP,MeMOTR} protocols), are shared across all benchmarks.
Only three settings track the benchmark: the training window and inference miss tolerance scale with typical occlusion length ($T{=}W{=}30$ frames for DanceTrack and PersonPath22, $60$ for SportsMOT); the ID dictionary size scales with crowd density ($K{=}300$ for the crowded PersonPath22, $50$ otherwise); and the detection and newborn thresholds $(\lambda_{\text{det}}, \lambda_{\text{newborn}})$ are $(0.4, 0.6)$ for PersonPath22 and $(0.3, 0.7)$ for DanceTrack and SportsMOT.

\subsection{Comparisons with State-of-the-art Methods}
\label{sec:sota}
We compare MOTIP2 with state-of-the-art methods on the DanceTrack, SportsMOT, and PersonPath22 test splits, each using the corresponding model described above.

\paragraph{DanceTrack.} DanceTrack stresses association under near-identical appearance and highly non-linear motion, the regime where learned association should help most (\cref{tab:sota_dancetrack}). Trained on DanceTrack alone, our default MOTIP2-L already reaches $\mathbf{73.4}$ HOTA, above the strongest prior end-to-end method (FDTA, $71.7$) and every tracking-by-detection tracker. The larger MOTIP2-X raises this to $74.8$ HOTA, $84.3$ DetA, $66.5$ AssA, and $78.7$ IDF1, the best of any method trained without extra data; its HOTA surpasses even the extra-data FDTA$^\dagger$ ($74.4$). Disabling all three priors while keeping every other component isolates their contribution: MOTIP2-L drops to $71.5$ HOTA, below FDTA, and MOTIP2-X to $72.7$, losing $3.2$ and $3.5$ AssA while DetA moves by at most $0.2$. The margin over prior end-to-end trackers is therefore associative, not an effect of the DEIM detector. Training additionally on the validation split (the $\dagger$ extra-data protocol) lifts both variants further, with MOTIP2-X$^\dagger$ setting a new overall state of the art at $\mathbf{77.1}$ HOTA, $85.4$ DetA, $69.8$ AssA, and $81.2$ IDF1 (MOTIP2-L$^\dagger$ reaches $76.1$ HOTA).

Both variants keep gaining under the extra-data protocol ($+2.3$ for MOTIP2-X, $+2.7$ for MOTIP2-L) rather than saturating. Across both regimes the gains are largest in association: MOTIP2-X lifts AssA by $+6.1$ over MOTIP and $+3.0$ over FDTA, while the detection gains track the stronger DEIM backbone.

\begin{table}[!tbp]
    \centering
    \setlength{\tabcolsep}{3pt}
    \small
    \begin{tabular}{lccccc}
        \toprule
        \textbf{Method} & \textbf{HOTA} & \textbf{DetA} & \textbf{AssA} & \textbf{MOTA} & \textbf{IDF1} \\
        \midrule
        \multicolumn{6}{l}{\textit{Tracking-by-detection}} \\
        FairMOT~\cite{FairMOT} & 39.7 & 66.7 & 23.8 & 82.2 & 40.8 \\
        CenterTrack~\cite{CenterTrack} & 41.8 & 78.1 & 22.6 & 86.8 & 35.7 \\
        ByteTrack~\cite{ByteTrack} & 47.7 & 71.0 & 32.1 & 89.6 & 53.9 \\
        OC-SORT~\cite{OC-SORT} & 55.1 & 80.3 & 38.3 & 92.0 & 54.6 \\
        StrongSORT~\cite{StrongSORT} & 55.6 & 80.7 & 38.6 & 91.1 & 55.2 \\
        C-BIoU~\cite{C-BIoU} & 60.6 & 81.3 & 45.4 & 91.6 & 61.6 \\
        C-TWiX~\cite{C-TWiX} & 62.1 & 81.8 & 47.2 & 91.4 & 63.6 \\
        DiffMOT~\cite{DiffMOT} & 62.3 & 82.5 & 47.2 & 92.8 & 63.0 \\
        UCMCTrack~\cite{UCMCTrack} & 63.6 & -- & -- & 88.9 & 65.0 \\
        FastTracker~\cite{FastTracker} & 65.9 & -- & -- & 93.4 & 67.2 \\
        \midrule
        \multicolumn{6}{l}{\textit{End-to-end}} \\
        MOTR~\cite{MOTR} & 54.2 & 73.5 & 40.2 & 79.7 & 51.5 \\
        FastTrackTr~\cite{FastTrackTr} & 62.4 & 78.4 & 49.8 & 88.8 & 64.8 \\
        CO-MOT~\cite{CO-MOT} & 65.3 & 80.1 & 53.5 & 89.3 & 66.5 \\
        MeMOTR~\cite{MeMOTR} & 68.5 & 80.5 & 58.4 & 89.9 & 71.2 \\
        MOTIP~\cite{MOTIP} & 69.6 & 80.4 & 60.4 & 90.6 & 74.7 \\
        FDTA~\cite{FDTA} & 71.7 & 81.0 & 63.5 & 91.3 & 77.2 \\
        \rowcolor{blueL!25} MOTIP2-L (w/o spatial priors) & 71.5 & 83.0 & 61.7 & 91.6 & 73.8 \\
        \rowcolor{blueL!25} MOTIP2-L & 73.4 & 83.1 & 64.9 & 91.6 & 77.0 \\
        \rowcolor{blueL!25} MOTIP2-X (w/o spatial priors) & 72.7 & 84.1 & 63.0 & 92.4 & 75.8 \\
        \rowcolor{blueL!25} MOTIP2-X & \textbf{74.8} & \textbf{84.3} & \textbf{66.5} & \textbf{92.5} & \textbf{78.7} \\
        \midrule
        \multicolumn{6}{l}{\textit{End-to-end with extra data}} \\
        CO-MOT$^\dagger$~\cite{CO-MOT} & 69.4 & 82.1 & 58.9 & 91.2 & 71.9 \\
        MOTRv2$^\dagger$~\cite{MOTRv2} & 69.9 & 83.0 & 59.0 & 91.9 & 71.7 \\
        MOTRv3$^\dagger$~\cite{MOTRv3} & 70.4 & 83.8 & 59.3 & 92.9 & 72.3 \\
        MOTIP$^\dagger$~\cite{MOTIP} & 72.0 & 81.8 & 63.5 & 91.9 & 76.8 \\
        FDTA$^\dagger$~\cite{FDTA} & 74.4 & 82.7 & 67.0 & 92.2 & 80.0 \\
        \rowcolor{blueL!25} MOTIP2-L$^\dagger$ & 76.1 & 84.9 & 68.4 & 92.9 & 79.6 \\
        \rowcolor{blueL!25} MOTIP2-X$^\dagger$ & \textbf{77.1} & \textbf{85.4} & \textbf{69.8} & \textbf{93.3} & \textbf{81.2} \\
        \bottomrule
    \end{tabular}
    \caption{\textbf{State-of-the-art comparison on the DanceTrack~\cite{DanceTrack} test set.} The best end-to-end result is in \textbf{bold}. $\dagger$~denotes extra training data (MOTIP2: DanceTrack train+val). MOTIP2-L/X (w/o spatial priors) disable all three priors under the same training setup, keeping every other component, which isolates their contribution from the DEIM detector.}
    \label{tab:sota_dancetrack}
\end{table}

\paragraph{SportsMOT.} SportsMOT couples fast non-linear motion with frequent occlusions in basketball, volleyball, and soccer, stressing association under abrupt direction changes rather than crowd density (\cref{tab:sota_sportsmot}). Trained on SportsMOT alone, MOTIP2-L is the best end-to-end tracker on every metric and the best overall on the identity-centric ones: $\mathbf{76.0}$ HOTA ($+1.8$ over FDTA, $+3.4$ over MOTIP, and $+3.9$ over the strongest tracking-by-detection method, DiffMOT~\cite{DiffMOT}), $85.6$ DetA, $67.5$ AssA, $93.8$ MOTA, and $78.9$ IDF1. As on DanceTrack, association accounts for most of the margin ($+7.0$ AssA over DiffMOT).

\begin{table}[!tbp]
    \centering
    \setlength{\tabcolsep}{3pt}
    \small
    \begin{tabular}{lccccc}
        \toprule
        \textbf{Method} & \textbf{HOTA} & \textbf{DetA} & \textbf{AssA} & \textbf{MOTA} & \textbf{IDF1} \\
        \midrule
        \multicolumn{6}{l}{\textit{Tracking-by-detection}} \\
        FairMOT~\cite{FairMOT} & 49.3 & 70.2 & 34.7 & 86.4 & 53.5 \\
        GTR~\cite{GTR} & 54.5 & 64.8 & 45.9 & 67.9 & 55.8 \\
        QDTrack~\cite{QDTrack} & 60.4 & 77.5 & 47.2 & 90.1 & 62.3 \\
        CenterTrack~\cite{CenterTrack} & 62.7 & 82.1 & 48.0 & 90.8 & 60.0 \\
        ByteTrack~\cite{ByteTrack} & 62.8 & 77.1 & 51.2 & 94.1 & 69.8 \\
        OC-SORT~\cite{OC-SORT} & 68.1 & 84.8 & 54.8 & 93.4 & 68.0 \\
        BoT-SORT~\cite{BoT-SORT} & 68.7 & 84.4 & 55.9 & 94.5 & 70.0 \\
        DiffMOT~\cite{DiffMOT} & 72.1 & 86.0 & 60.5 & 94.5 & 72.8 \\
        \midrule
        \multicolumn{6}{l}{\textit{End-to-end}} \\
        TransTrack~\cite{TransTrack} & 68.9 & 82.7 & 57.5 & 92.6 & 71.5 \\
        SambaMOTR~\cite{SambaMOTR} & 69.8 & 82.2 & 59.4 & 90.3 & 71.9 \\
        MeMOTR~\cite{MeMOTR} & 70.0 & 83.1 & 59.1 & 91.5 & 71.4 \\
        MOTIP~\cite{MOTIP} & 72.6 & 83.5 & 63.2 & 92.4 & 77.1 \\
        FDTA~\cite{FDTA} & 74.2 & 84.1 & 65.5 & 93.0 & 78.5 \\
        \rowcolor{blueL!25} MOTIP2-L & \textbf{76.0} & \textbf{85.6} & \textbf{67.5} & \textbf{93.8} & \textbf{78.9} \\
        \bottomrule
    \end{tabular}
    \caption{\textbf{State-of-the-art comparison on the SportsMOT~\cite{SportsMOT} test set.} The best end-to-end result is in \textbf{bold}.}
    \label{tab:sota_sportsmot}
\end{table}

\paragraph{PersonPath22.} PersonPath22 is a large-scale real-world benchmark ($236$ videos spanning diverse indoor and outdoor scenes, lighting and weather conditions, camera viewpoints, and crowd densities) targeting smart-home and smart-city deployment (\cref{tab:sota_personpath22}). As the only modern end-to-end tracker reported on it, MOTIP2-L improves substantially over TrackFormer (IDF1 $\mathbf{71.1}$ vs $57.1$, MOTA $75.8$ vs $69.7$) and attains the best IDF1 overall, $+1.7$ over the strongest dedicated tracker, Path Consistency~\cite{PathConsistency} ($69.4$), while matching its MOTA ($75.8$). The count-based metrics show the expected trade-off of identity-centric association: MOTIP2-L keeps identity switches competitive ($5\,045$, fewer than ByteTrack and TrackFormer) but produces more false positives than the most conservative trackers. Prior work here does not report HOTA-family metrics; for reference, MOTIP2-L reaches $62.8$ HOTA ($66.1$ DetA, $60.4$ AssA).

\begin{table}[!tbp]
    \centering
    \setlength{\tabcolsep}{3pt}
    \small
    \begin{tabular}{lccccc}
        \toprule
        \textbf{Method} & \textbf{IDF1} & \textbf{MOTA} & \textbf{IDSW}$\downarrow$ & \textbf{FP}$\downarrow$ & \textbf{FN}$\downarrow$ \\
        \midrule
        \multicolumn{6}{l}{\textit{Tracking-by-detection}} \\
        CenterTrack~\cite{CenterTrack} & 46.4 & 59.3 & 10\,319 & 24K & 72K \\
        FairMOT~\cite{FairMOT} & 61.1 & 61.8 & 5\,095 & 15K & 80K \\
        ByteTrack~\cite{ByteTrack} & 66.8 & 75.4 & 5\,931 & 17K & 40K \\
        OC-SORT~\cite{OC-SORT} & 68.2 & 74.0 & 4\,931 & 17K & 41K \\
        Path Consistency~\cite{PathConsistency} & 69.4 & 75.8 & 4\,663 & 15K & 38K \\
        \midrule
        \multicolumn{6}{l}{\textit{End-to-end}} \\
        TrackFormer~\cite{TrackFormer} & 57.1 & 69.7 & 8\,633 & 23K & 47K \\
        \rowcolor{blueL!25} MOTIP2-L & \textbf{71.1} & \textbf{75.8} & \textbf{5\,045} & \textbf{18K} & \textbf{40K} \\
        \bottomrule
    \end{tabular}
    \caption{\textbf{State-of-the-art comparison on the PersonPath22~\cite{PersonPath22} test set.} The best end-to-end result is in \textbf{bold}. Numbers for comparison methods are compiled from~\cite{PersonPath22} and~\cite{PathConsistency}; these works do not report HOTA-family metrics. For completeness, MOTIP2-L additionally achieves HOTA 62.8, DetA 66.1, and AssA 60.4.}
    \label{tab:sota_personpath22}
\end{table}

\subsection{Speed and Accuracy}
\label{sec:speed}

Trackers typically run online, on a live video stream feeding downstream tasks, and must keep pace with the incoming frames; we therefore evaluate accuracy jointly with latency.

\Cref{fig:hota_fps} plots the HOTA-versus-latency trade-off on DanceTrack-val (the state-of-the-art tables above report the test split), with the exact per-model HOTA and latency behind each point listed in \cref{sec:suppl_latency}. Across model sizes, MOTIP2 traces a Pareto front ahead of prior end-to-end trackers (FastTrackTr~\cite{FastTrackTr}, PuTR~\cite{PuTR}): MOTIP2-S matches MOTIP's accuracy while running over $\mathbf{3\times}$ faster ($27.1$ vs $86.9$ ms), and our larger variants surpass them in accuracy while remaining faster. The association advantage holds even before detection catches up: with a shallower ID decoder than MOTIP ($3$ vs $6$ layers), MOTIP2-S surpasses MOTIP's HOTA despite $1.4$ lower DetA ($+1.8$ AssA), and once detection reaches MOTIP's level (MOTIP2-M) association leads by $+5.0$ AssA. The improvement is concentrated in association, not the DEIM backbone.

\paragraph{Latency breakdown.}
\Cref{tab:latency_trt} reports per-component latency for MOTIP2 across V100 (PyTorch FP32) and T4 (TensorRT FP16), profiling DEIM and the ID decoder separately; a full comparison against baselines is given in \cref{sec:suppl_latency}. DEIM latency is fixed once the model size is chosen, while the ID decoder's latency grows roughly linearly with the number of tracked objects in the trajectory history. Sparser scenes such as DanceTrack therefore run faster, while denser scenes incur only a moderate increase. MOTIP has no entry in the TensorRT columns: its Deformable DETR detector relies on deformable attention, which has no standard TensorRT path, so it cannot be exported or benchmarked on a deployment GPU. Both MOTIP2 components export to TensorRT as separate engines, where MOTIP2-S reaches $10.1$ ms ($\approx 99$ FPS) on a T4 under FP16.

\begin{table*}[!tbp]
    \centering
    \setlength{\tabcolsep}{3pt}
    \small
    \begin{tabular}{l @{\hskip 12pt} ccc @{\hskip 12pt} ccc @{\hskip 12pt} ccc}
        \toprule
        & \multicolumn{3}{c}{\textbf{Accuracy}} & \multicolumn{3}{c}{\textbf{V100 PT FP32}} & \multicolumn{3}{c}{\textbf{T4 TRT FP16}} \\
        \cmidrule(lr){2-4} \cmidrule(lr){5-7} \cmidrule(lr){8-10}
        \textbf{Method} & \textbf{HOTA} & \textbf{DetA} & \textbf{AssA} & \textbf{Det.} & \textbf{ID Dec} & \textbf{Total} & \textbf{Det.} & \textbf{ID Dec} & \textbf{Total} \\
        \midrule
        MOTIP~\cite{MOTIP}   & 64.0 & 75.4 & 54.7 & 72.7 & 14.2 & 86.9 & \multicolumn{3}{c}{--} \\
        \midrule
        \rowcolor{blueL!25} MOTIP2-X & 70.1 & 80.7 & 61.1 & 71.2 & 9.9 & 81.1 & 38.8 & 1.3 & 40.1 \\
        \rowcolor{blueL!25} MOTIP2-L & 68.0 & 77.3 & 60.0 & 45.0 & 9.8 & 54.8 & 21.9 & 1.3 & 23.2 \\
        \rowcolor{blueL!25} MOTIP2-M & 67.3 & 76.4 & 59.7 & 32.5 & 6.8 & 39.3 & 15.8 & 0.9 & 16.7 \\
        \rowcolor{blueL!25} MOTIP2-S & 64.5 & 74.0 & 56.5 & 21.8 & 5.3 & 27.1 & 9.4 & 0.7 & 10.1 \\
        \rowcolor{blueL!25} MOTIP2-N & 59.2 & 69.3 & 50.9 & 14.9 & 3.9 & 18.8 & 4.0 & 0.4 & 4.4 \\
        \bottomrule
    \end{tabular}
    \caption{\textbf{Per-component latency breakdown.} HOTA on DanceTrack-val and per-component latency (ms). Det.\ = detector (Deformable DETR for MOTIP, DEIM for MOTIP2); PT = PyTorch, TRT = TensorRT. MOTIP has no standard TensorRT path. Input resolution: $800\times1440$. ID Decoder profiled with $T{=}30$ frames, 50 trajectories, and 50 detections.}
    \label{tab:latency_trt}
\end{table*}

\subsection{Ablation Studies}
\label{sec:ablations}

All ablations use MOTIP2-S. We train one model per evaluation benchmark following its standard recipe: one on PersonPath22+CrowdHuman, scored on the PersonPath22 test set, and one on the DanceTrack training split, scored on the DanceTrack validation set. Every ablated variant is retrained under both recipes and evaluated on its corresponding split.

\begin{table*}[!tbp]
    \centering
    \small
    \setlength{\tabcolsep}{6pt}
    \begin{tabular}{@{}l@{\quad}cccc@{\qquad}cccc@{}}
        \toprule
        & \multicolumn{4}{c}{\textbf{PersonPath22}} & \multicolumn{4}{c}{\textbf{DanceTrack}} \\
        \cmidrule(lr){2-5} \cmidrule(lr){6-9}
        & \textbf{HOTA} & \textbf{DetA} & \textbf{AssA} & \textbf{IDF1}
        & \textbf{HOTA} & \textbf{DetA} & \textbf{AssA} & \textbf{IDF1} \\
        \midrule
        \textbf{Full model}                          & 61.6 & 63.8 & 60.3 & 70.4 & 64.5 & 74.0 & 56.5 & 68.8 \\
        \midrule
        uniform IDSW                                 & 56.1 & 58.6 & 54.4 & 64.0 & 63.7 & 73.0 & 55.8 & 68.1 \\
        w/o IDSW                                     & 59.5 & 64.2 & 55.8 & 66.8 & 60.1 & 73.3 & 49.7 & 63.0 \\
        \midrule
        w/o Spatial ID Loss                          & 61.4 & 64.2 & 59.4 & 70.1 & 63.5 & 73.3 & 55.5 & 67.3 \\
        w/o Spatial Anchor                           & 61.6 & 64.3 & 59.9 & 70.4 & 64.0 & 73.3 & 56.1 & 68.6 \\
        \midrule
        w/o all priors\textsuperscript{\dag}         & 58.7 & 64.1 & 54.5 & 65.7 & 63.0 & 74.0 & 54.0 & 66.9 \\
        \bottomrule
    \end{tabular}
    \caption{\textbf{Ablation of the spatial priors.} MOTIP2-S, each variant trained and evaluated per benchmark (PersonPath22+CrowdHuman; DanceTrack train split). \emph{uniform IDSW} downgrades our Spatial ID Switches to a non-spatial permutation; \emph{w/o IDSW} removes the augmentation entirely. \textsuperscript{\dag}With all priors off, the data-stage prior is removed on PersonPath22 but set to uniform on DanceTrack, where uniform is near-neutral rather than harmful (cf.\ rows above). The full model (bold row label) is the reference each removal is compared against.}
    \label{tab:ablation_removal}
\end{table*}

The three spatial priors all inject geometric structure into association (\cref{tab:ablation_removal}). Removing all three at once costs $-2.9$ HOTA and $-5.8$ AssA on PersonPath22 and $-1.5$ HOTA and $-2.5$ AssA on DanceTrack; \cref{sec:suppl_longswitch} shows the qualitative effect on a DanceTrack sequence, where the no-prior baseline teleports an identity across the frame. The loss- and representation-stage priors hold up even with the data-stage prior neutralized: against the matched reference (\emph{w/o IDSW} on PersonPath22, \emph{uniform IDSW} on DanceTrack, both of which keep the loss and anchor under the same data-stage setting), dropping them costs a further $-0.8$ / $-0.7$ HOTA and $-1.3$ / $-1.8$ AssA, so their benefit is not merely an artifact of the spatial augmentation. The two benchmarks expose them differently: PersonPath22's wide, crowded scenes make the data-stage prior decisive, while DanceTrack's compact crowds of near-identical dancers bring out the loss- and representation-stage priors. We discuss each in turn, in the order of \cref{sec:method}, then ablate two architectural tweaks to the ID decoder and inference separately (\cref{sec:arch_ablations}).

\begin{figure*}[t]
  \centering
  \begin{tikzpicture}
    \node[anchor=south west, inner sep=0] (grid)
      {\includegraphics[width=\dimexpr\textwidth-11pt\relax]{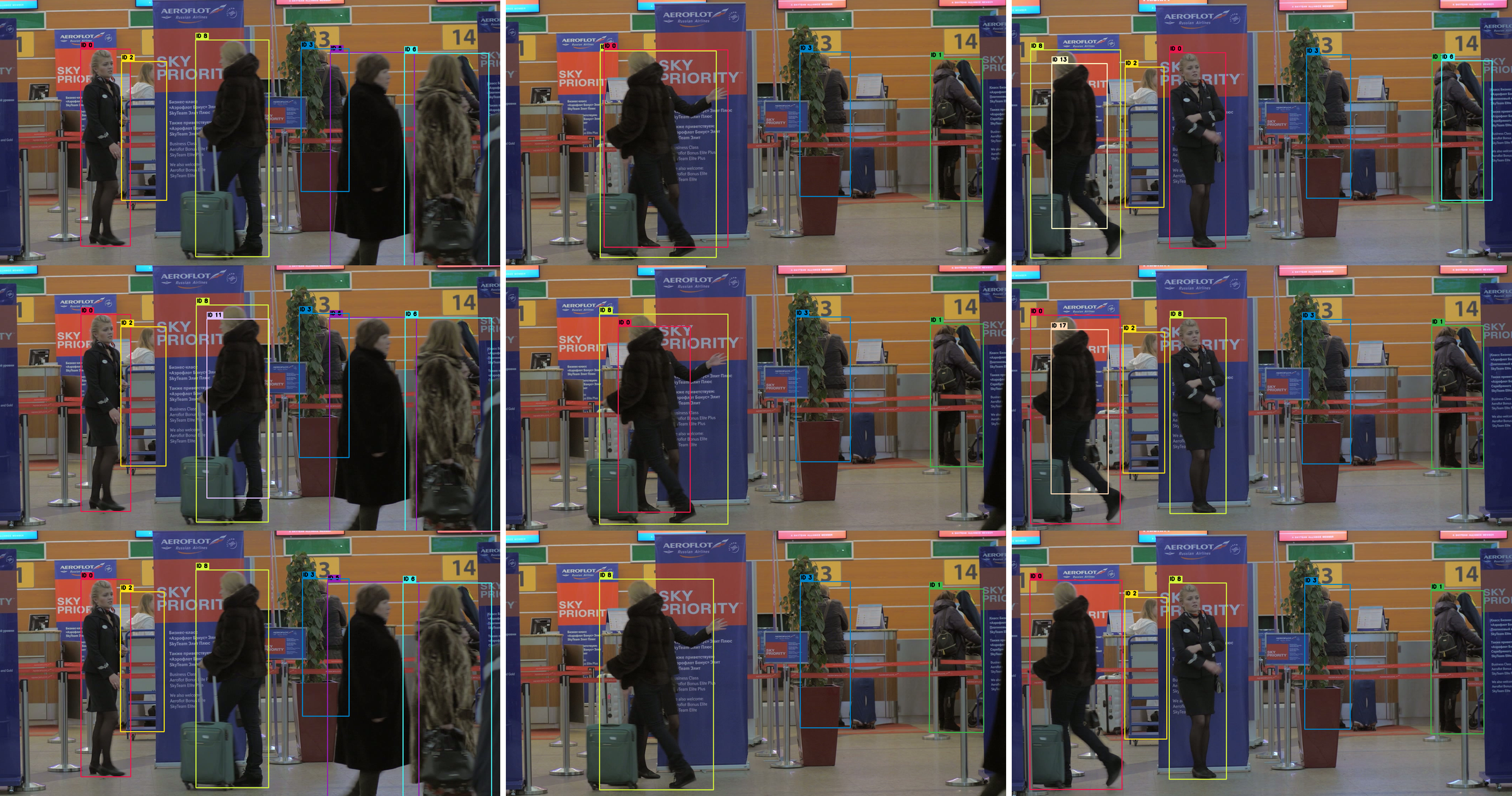}};
    \begin{scope}[x={(grid.south east)}, y={(grid.north west)}]
      \node[rotate=90, anchor=south, inner sep=3pt, font=\footnotesize] at (0, 0.8333) {Ours};
      \node[rotate=90, anchor=south, inner sep=3pt, font=\footnotesize] at (0, 0.5) {w/o IDSW};
      \node[rotate=90, anchor=south, inner sep=3pt, font=\footnotesize] at (0, 0.1667) {uniform IDSW};
    \end{scope}
  \end{tikzpicture}
  \caption{\textbf{Qualitative tracking comparison on PersonPath22}
  (uid\_vid\_00063). Rows (top to bottom): our GIoU-based sampling
  (\emph{Ours}), no augmentation (\emph{w/o IDSW}), and MOTIP's uniform
  permutation (\emph{uniform IDSW}). Columns are ordered in time, left to
  right. Two pedestrians cross and occlude; only ours recovers the correct
  identities afterward.}
  \label{fig:idsw_qual}
\end{figure*}
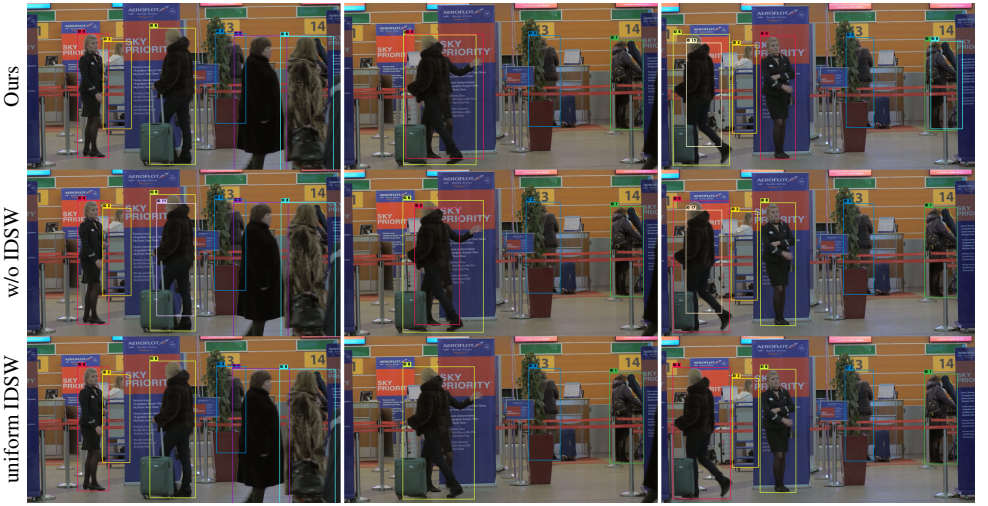

\paragraph{Spatial ID Switches.}
Spatial ID Switches are by far the most important prior on PersonPath22, where they also expose a failure mode of the uniform permutation they replace. Removing the augmentation altogether costs $-2.1$ HOTA (almost all association, $-4.5$ AssA), but reverting it to MOTIP's uniform permutation is worse still: at $56.1$ HOTA it falls $3.4$ points \emph{below} the no-augmentation baseline ($59.5$) and drags detection down with it ($-5.6$ DetA). On PersonPath22 a uniform permutation is therefore harmful, not just uninformative: it destabilizes joint training instead of teaching the decoder to recover from confusions (training curves in \cref{sec:suppl_idsw_stability}). \Cref{fig:idsw_qual} shows the three behaviors on a representative crossing: only our model recovers the correct identities, the no-augmentation variant trusts its trajectory history blindly and never undoes the swap, and the uniform variant grows uncertain through the occlusion, losing the detection before switching identity.

The cause is the train/inference mismatch that GIoU sampling removes (\cref{sec:id_switches}): PersonPath22 spreads objects across the frame, so uniform permutations fabricate mostly implausible long-distance swaps. Restricting confusions to spatially plausible pairs recovers $\mathbf{+5.5}$ HOTA over the uniform variant. Training on the confusions the tracker meets at inference concentrates the decoder's uncertainty on the few tracks an object actually crosses, rather than on every identity in the frame. DanceTrack shows why the effect is dataset-dependent: the augmentation is again critical ($-4.4$ HOTA when removed), but its uniform variant costs only $-0.8$, because dancers stay clustered, so even a random permutation swaps mostly nearby objects.

\paragraph{Spatial ID Loss.}
The Spatial ID Loss (\cref{sec:spatial_loss}) mainly improves association and complements the dominant Spatial ID Switches: removing it costs only $-0.2$ HOTA and $-0.9$ AssA on PersonPath22, but $-1.0$ HOTA and $-1.0$ AssA on DanceTrack, where near-identical appearance makes geometric supervision more valuable.

\paragraph{Spatial Anchor.}
Reinjecting the detector's per-frame positional encoding into every ID-decoder token is the lightest of the three priors: no new parameters, no new supervision, just reusing an encoding the detector has already produced. We expect a correspondingly modest gain, concentrated where many candidate tracks share similar content embeddings and an explicit spatial cue is most discriminative. The data agree: removal costs $-0.4$ AssA and no HOTA on PersonPath22, but $-0.5$ HOTA and $-0.4$ AssA on DanceTrack, where appearance alone is weakly discriminative.

\subsection{Decoder and Inference Ablations}
\label{sec:arch_ablations}

Two changes to MOTIP2's ID decoder and inference fall outside the spatial priors: rotary temporal encoding in the decoder and tracklet confirmation at inference.

\paragraph{Rotary Position Embedding.}
Replacing MOTIP's learned temporal bias with RoPE lifts AssA by $+1.0$ and HOTA by $+0.3$ at the best timescale (\cref{tab:ablation_rope_vs_rel_pe}). The gain hinges on the inference timescale $\alpha$: training samples wider temporal intervals than inference, so the raw encoding ($\alpha{=}1.0$) trails the learned bias, while rescaling toward the training mean recovers and surpasses it, plateauing across $\alpha \in [2.0, 2.5]$. We adopt $\alpha{=}2.0$.

\begin{table}[!tbp]
    \centering
    \setlength{\tabcolsep}{6pt}
    \small
    \begin{tabular}{@{}l c ccccc@{}}
        \toprule
        \textbf{Method} & $\alpha$ & \textbf{HOTA} & \textbf{DetA} & \textbf{AssA} & \textbf{MOTA} & \textbf{IDF1} \\
        \midrule
        Rel.\ PE & -- & 61.3 & \textbf{64.1} & 59.3 & \textbf{73.4} & 70.2 \\
        \midrule
        \multirow{4}{*}{RoPE}
        & 1.0 & 60.4 & 63.7 & 58.0 & 72.7 & 68.7 \\
        & 1.5 & 61.0 & 63.7 & 59.1 & 72.8 & 69.6 \\
        & 2.0 & \textbf{61.6} & 63.8 & \textbf{60.3} & 72.9 & \textbf{70.4} \\
        & 2.5 & \textbf{61.6} & 63.8 & \textbf{60.3} & 72.9 & \textbf{70.4} \\
        \bottomrule
    \end{tabular}
    \caption{\textbf{Effect of positional encoding type and time rescaling factor $\alpha$.} Results on PersonPath22 test. We compare RoPE with the learned relative positional encoding from MOTIP~\cite{MOTIP} (Rel.\ PE). The best result for each metric is shown in \textbf{bold}.}
    \label{tab:ablation_rope_vs_rel_pe}
\end{table}

\paragraph{Tracklet Confirmation.}
Requiring $N_{\text{init}}$ consecutive frames of consistent assignment before committing a tentative track trims identity switches without hurting accuracy: from $N_{\text{init}}{=}1$ to $3$, identity switches drop by $12.5\%$ on PersonPath22 and $28.3\%$ on DanceTrack, while HOTA stays flat and IDF1 and MOTA edge up only slightly. Evicting unconfirmed tracks also keeps the historical context clean, freeing ID-bank slots that short-lived false detections would otherwise occupy. Accuracy plateaus at $N_{\text{init}}{=}2$, and identity switches fall only marginally beyond it, so we adopt $N_{\text{init}}{=}2$; the full sweep is in \cref{sec:suppl_confirmation}.

\section{Conclusion}
\label{sec:conclusion}

We argue that the association errors separating end-to-end trackers from tracking-by-detection are mainly geometric, and we correct them with three explicit spatial priors. Which prior helps most depends on scene geometry rather than the benchmark. In wide, crowded scenes the data-stage prior is decisive: the uniform shuffling it replaces trains the model on implausible long-distance swaps and makes training unstable, whereas restricting confusions to nearby boxes recovers $\mathbf{+5.5}$ HOTA on PersonPath22. When objects look nearly identical, the other two matter more. The loss-stage prior makes a confident long-distance error cost more than a nearby one, and the representation-stage prior ties each track to its image position; both separate identities that appearance alone confuses. This reliance on proximity also sets the failure case we expect to be hardest: a simultaneous exit and entry at the same location, where proximity draws the leaving identity onto the new object. The result is a family of end-to-end trackers spanning the speed-accuracy trade-off, from a real-time model to variants that set a new state of the art on identity association across all three benchmarks.

\section*{Acknowledgements}
This work was granted access to the HPC resources of IDRIS under the allocations AD011014610 and A0191013894 made by GENCI.

\bibliography{main}

\clearpage
\appendix
\makeatletter
\newcommand{\maketitlesupplementary}{%
  \begin{center}
    \Large
    \textbf{\@title}\\
    \vspace{0.5em}Supplementary Material\\
    \vspace{1.0em}
  \end{center}
}
\makeatother
\clearpage
\maketitlesupplementary

\section{Training Details}
\label[appendix]{sec:suppl_training}

\paragraph{Optimizer Configuration.}
We use AdamW~\cite{AdamW} with per-parameter-group learning rates and weight decay. The encoder and decoder use a base learning rate of $8\times10^{-4}$ for MOTIP2-N, $2\times10^{-4}$ for MOTIP2-S/M, and $2.5\times10^{-4}$ for MOTIP2-L/X. The ID decoder uses twice this rate. The backbone uses $0.5\times$ this rate for MOTIP2-N/S, $0.1\times$ for MOTIP2-M, $0.05\times$ for MOTIP2-L, and $0.01\times$ for MOTIP2-X. Weight decay is set per model size as well and is disabled for normalization layers, bias terms, and embedding parameters. Gradients are clipped to a max norm of $0.1$. We apply EMA with a decay of $0.9995$, with no warmup during pretraining and a 1000-step warmup during joint training.

\paragraph{Learning Rate Schedule.}
Training follows a two-stage schedule, both stages at a constant learning rate with 40k samples per epoch. We first pretrain the DEIM~\cite{DEIM} detector alone, with no learning-rate warmup, then jointly train the full model (detector + ID decoder) with a 500-iteration linear warmup.

\paragraph{Batch Size.}
For MOTIP2-S/M/L/X, we use 8 images per GPU during detector pretraining and 2 clips per GPU during joint training, giving global batch sizes of 32 images and 8 clips across four GPUs. MOTIP2-N uses 32 images and 4 clips per GPU, giving global batch sizes of 128 images and 16 clips, respectively. We do not accumulate gradients.

\paragraph{Clip Sampling.}
We sample clips of 30 frames (60 for SportsMOT) with a temporal interval uniformly sampled from 1 to 4. The detection loss is computed on 4 randomly sampled frames per clip (8 for the nano model), while the ID decoder receives the full clip.

\paragraph{Data Augmentation Details.}
We apply augmentations at two levels: image-level and trajectory-level.

\textit{Image-level geometric.} Horizontal flip with probability $0.5$; random zoom-out with probability $0.5$ and scale range $[1.0, 2.0]$; IoU-aware random crop with probability $0.8$, scale range $[0.3, 1.0]$, and aspect ratio range $[1.5, 2.0]$.

\textit{Multi-scale resizing.} Every clip is randomly resized to one of a set of resolutions spanning roughly $75\%$ to $125\%$ of the base size, each snapped to a multiple of $32$, with a tunable sampling bias toward the base resolution; all frames within a clip share the sampled scale. Exposing the model to many resolutions during training yields a single network that handles a range of input sizes at inference without retraining, so the test-time resolution becomes a free accuracy-latency dial.

\textit{Photometric.} Applied with probability $0.5$: brightness jitter in $[0.875, 1.125]$, contrast in $[0.5, 1.5]$, saturation in $[0.5, 1.5]$, and hue shift in $[-0.05, 0.05]$.

\textit{Trajectory-level.} Spatial ID Switches are applied with probability $1.0$, sampling identity permutations weighted by pairwise GIoU affinity. Random occlusion masking is applied with probability $0.5$. Trajectories are organized into 6 groups of 100 IDs each.

\paragraph{Loss Weights.}
The DEIM detection losses use the following weights: classification (MAL) $1.0$, bounding box L1 $5.0$, GIoU $2.0$, FGL $0.15$, and DDF $1.5$ (the last two from D-FINE~\cite{D-FINE}), with focal loss~\cite{FocalLoss} parameters $\alpha{=}0.75$ and $\gamma{=}1.5$. The Hungarian matcher costs are: class $2.0$, bounding box $5.0$, GIoU $2.0$. The ID association loss weight is $1.0$. Following quality-aware detection losses~\cite{VFL,QFL}, the ID cross-entropy is weighted per-sample by the predicted-to-ground-truth IoU, concentrating association learning on well-localized detections.

\section{Reference Comparison on MOT17}
\label[appendix]{sec:suppl_mot17}

We train MOTIP2-L/X on MOT17~\cite{MOT17} and CrowdHuman~\cite{CrowdHuman}, matching MOTIP's recipe, and evaluate on the MOT17 test set (\cref{tab:sota_mot17}). MOTIP2 improves over MOTIP~\cite{MOTIP} on every metric, including association ($+1.0$ AssA and $+1.4$ IDF1 at L), though the gains concentrate in detection ($+1.9$ DetA at X). Scaling L to X leaves AssA flat at $58.0$: association on MOT17 is data-bound rather than capacity-bound, since CrowdHuman's static images supervise detection only, leaving MOT17's 7 training sequences as the sole association supervision, whose near-linear motion favors training-free Kalman~\cite{KalmanFilter} association. This regime motivated our focus on PersonPath22, a pedestrian benchmark with 138 training videos.

\begin{table}[t]
    \centering
    \setlength{\tabcolsep}{3pt}
    \small
    \begin{tabular}{lccccc}
        \toprule
        \textbf{Method} & \textbf{HOTA} & \textbf{DetA} & \textbf{AssA} & \textbf{MOTA} & \textbf{IDF1} \\
        \midrule
        \multicolumn{6}{l}{\textit{Tracking-by-detection}} \\
        CenterTrack~\cite{CenterTrack} & 52.2 & 53.8 & 51.0 & 67.8 & 64.7 \\
        FairMOT~\cite{FairMOT} & 59.3 & 60.9 & 58.0 & 73.7 & 72.3 \\
        DeepSORT~\cite{DeepSORT} & 61.2 & 63.1 & 59.7 & 78.0 & 74.5 \\
        SORT~\cite{SORT} & 63.0 & 64.2 & 62.2 & 80.1 & 78.2 \\
        ByteTrack~\cite{ByteTrack} & 63.1 & 64.5 & 62.0 & 80.3 & 77.3 \\
        OC-SORT~\cite{OC-SORT} & 63.2 & 63.2 & 63.4 & 78.0 & 77.5 \\
        C-BIoU~\cite{C-BIoU} & 64.1 & 64.8 & 63.7 & 81.1 & 79.7 \\
        MotionTrack~\cite{MotionTrack} & 65.1 & 65.4 & 65.1 & 81.1 & 80.1 \\
        SparseTrack~\cite{SparseTrack} & 65.1 & -- & -- & 81.0 & 80.1 \\
        FastTracker~\cite{FastTracker} & 66.4 & 66.3 & 66.9 & 81.8 & 82.0 \\
        BoostTrack++~\cite{BoostTrack++} & 66.6 & -- & -- & 80.7 & 82.2 \\
        \midrule
        \multicolumn{6}{l}{\textit{End-to-end}} \\
        TrackFormer~\cite{TrackFormer} & -- & -- & -- & 74.1 & 68.0 \\
        MOTR~\cite{MOTR} & 57.2 & 58.9 & 55.8 & 71.9 & 68.4 \\
        MOTRv2~\cite{MOTRv2} & 57.6 & 58.1 & 57.5 & 70.1 & 70.3 \\
        MeMOTR~\cite{MeMOTR} & 58.8 & 59.6 & 58.4 & 72.8 & 71.5 \\
        MOTIP~\cite{MOTIP} & 59.3 & 62.0 & 57.0 & 75.3 & 71.3 \\
        CO-MOT~\cite{CO-MOT} & 60.1 & 59.5 & 60.6 & 72.6 & 72.7 \\
        MOTRv3~\cite{MOTRv3} & 60.2 & 62.1 & 58.7 & 75.9 & 72.4 \\
        FastTrackTr~\cite{FastTrackTr} & \textbf{62.4} & 62.8 & \textbf{63.0} & 76.7 & \textbf{77.2} \\
        \rowcolor{blueL!25} MOTIP2-L & 60.4 & 63.2 & 58.0 & 76.3 & 72.7 \\
        \rowcolor{blueL!25} MOTIP2-X & 60.7 & \textbf{63.9} & 58.0 & \textbf{77.3} & 72.6 \\
        \bottomrule
    \end{tabular}
    \caption{\textbf{State-of-the-art comparison on the MOT17~\cite{MOT17} test set.} The best end-to-end result is in \textbf{bold}. MOTIP2-L/X are trained on MOT17 + CrowdHuman, matching MOTIP's recipe.}
    \label{tab:sota_mot17}
\end{table}

\section{Training Stability of Spatial ID Switches}
\label[appendix]{sec:suppl_idsw_stability}

\begin{figure}[!ht]
  \centering
  \begin{tikzpicture}
  \begin{axis}[
      width=0.8\linewidth, height=4.4cm,
      xlabel={Training epoch},
      ylabel={HOTA (\%)},
      xmin=-0.3, xmax=9.3,
      ymin=44, ymax=62,
      xtick={0,1,2,3,4,5,6,7,8,9},
      ytick={45,48,51,54,57,60},
      tick align=outside,
      grid=major, grid style={dashed, gray!30},
      legend cell align=left,
      legend columns=1,
      legend style={at={(0.97,0.5)}, anchor=east, draw=gray!50,
          fill=white, fill opacity=0.85, text opacity=1,
          font=\footnotesize},
      label style={font=\small},
      tick label style={font=\footnotesize},
      every axis plot/.append style={line width=0.9pt, mark size=1.6pt},
  ]
  \addplot[color=cOurs, mark=*] coordinates {
      (0,57.927) (1,59.022) (2,59.618) (3,59.971) (4,60.505)
      (5,60.872) (6,61.066) (7,60.958) (8,61.171) (9,61.419)
  };
  \addlegendentry{Ours (GIoU sampling)}
  \addplot[color=cNone, mark=triangle*, dashed] coordinates {
      (0,57.101) (1,57.571) (2,58.018) (3,58.566) (4,58.169)
      (5,58.389) (6,59.198) (7,58.867) (8,59.620) (9,59.069)
  };
  \addlegendentry{w/o IDSW}
  \addplot[color=cNaive, mark=square*] coordinates {
      (0,45.597) (1,53.341) (2,55.040) (3,56.106) (4,54.336)
      (5,52.552) (6,48.692) (8,48.154) (9,47.282)
  };
  \addlegendentry{uniform IDSW}
  \end{axis}
  \end{tikzpicture}
  \caption{\textbf{ID-switch augmentation variants.} PersonPath22 (test) HOTA
  over joint-training epochs (MOTIP2-S): GIoU-based sampling (\emph{Ours}), no
  augmentation (\emph{w/o IDSW}), and MOTIP's uniform permutation
  (\emph{uniform IDSW}). The uniform permutation diverges and ends below the
  no-augmentation baseline, while ours trains stably to the highest HOTA.}
  \label{fig:idsw_hota_per_epoch}
  \end{figure}
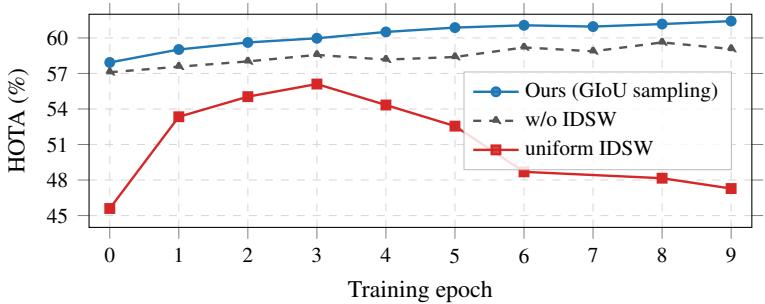

\Cref{fig:idsw_hota_per_epoch} tracks PersonPath22 HOTA across the joint-training epochs for the three augmentation variants ablated in \cref{sec:ablations}. The uniform permutation of MOTIP~\cite{MOTIP} improves for the first few epochs, then diverges and ends well below the no-augmentation baseline. Because PersonPath22's wide scenes spread objects across the frame, a uniform permutation fabricates mostly long-distance swaps that rarely occur at inference (\cref{sec:id_switches}); supervised against these implausible histories, the ID decoder destabilizes the jointly trained detector rather than learning to recover from realistic confusions. GIoU-based sampling (ours) restricts confusions to spatially plausible pairs, trains stably throughout, and reaches the best HOTA.

\Cref{tab:swap_stats} quantifies the augmentation over one epoch. The uniform permutation mostly swaps non-overlapping boxes (GIoU $\le 0$), fabricating long-range swaps; Spatial ID Switches realize fewer swaps, focused on overlapping objects (mean GIoU $0.20$/$0.15$), yet reach a higher HOTA. At inference, the uniform variant produces the most long-range ID switches ($2.2\times$/$1.7\times$ the no-augmentation counts), whereas ours stays close to the no-augmentation level while learning to recover from identity errors.

\begin{table}[t]
    \centering
    \small
    \setlength{\tabcolsep}{3pt}
    \begin{tabular}{lccc@{\quad}cc}
        \toprule
        & \multicolumn{3}{c@{\quad}}{\textbf{Training}} & \multicolumn{2}{c}{\textbf{Inference}} \\
        \cmidrule(r{1em}){2-4}\cmidrule(l){5-6}
        \textbf{Sampling} & \textbf{Swap\%} & \textbf{LR\%} & {\boldmath$\overline{\mathrm{GIoU}}$} & \textbf{HOTA} & \textbf{LR-sw} \\
        \midrule
        \multicolumn{6}{@{}l}{\emph{DanceTrack}} \\
        None & -- & -- & -- & 60.1 & 120 \\
        Uniform & 37\% & 85\% & $-0.40$ & 63.7 & 260 \\
        \rowcolor{blueL!25} Ours & 10\% & 26\% & $0.20$ & \textbf{64.5} & 134 \\
        \midrule
        \multicolumn{6}{@{}l}{\emph{PersonPath22}} \\
        None & -- & -- & -- & 59.5 & 186 \\
        Uniform & 43\% & 94\% & $-0.65$ & 56.1 & 322 \\
        \rowcolor{blueL!25} Ours & 10\% & 31\% & $0.15$ & \textbf{61.6} & 195 \\
        \bottomrule
    \end{tabular}
    \caption{\textbf{Spatial ID-switch augmentation behavior.} \emph{Training}, over one epoch: the realized swap rate (Swap\%), the fraction of swapped pairs that are long-range (LR\%: GIoU~$\le 0$, non-overlapping), and their mean overlap ($\overline{\mathrm{GIoU}}$). \emph{Inference}: the number of long-range ID switches produced by each variant (LR-sw).}
    \label{tab:swap_stats}
\end{table}

\section{A Long-Distance Identity Switch}
\label[appendix]{sec:suppl_longswitch}

\begin{figure*}[t]
  \centering
  \includegraphics[width=\textwidth]{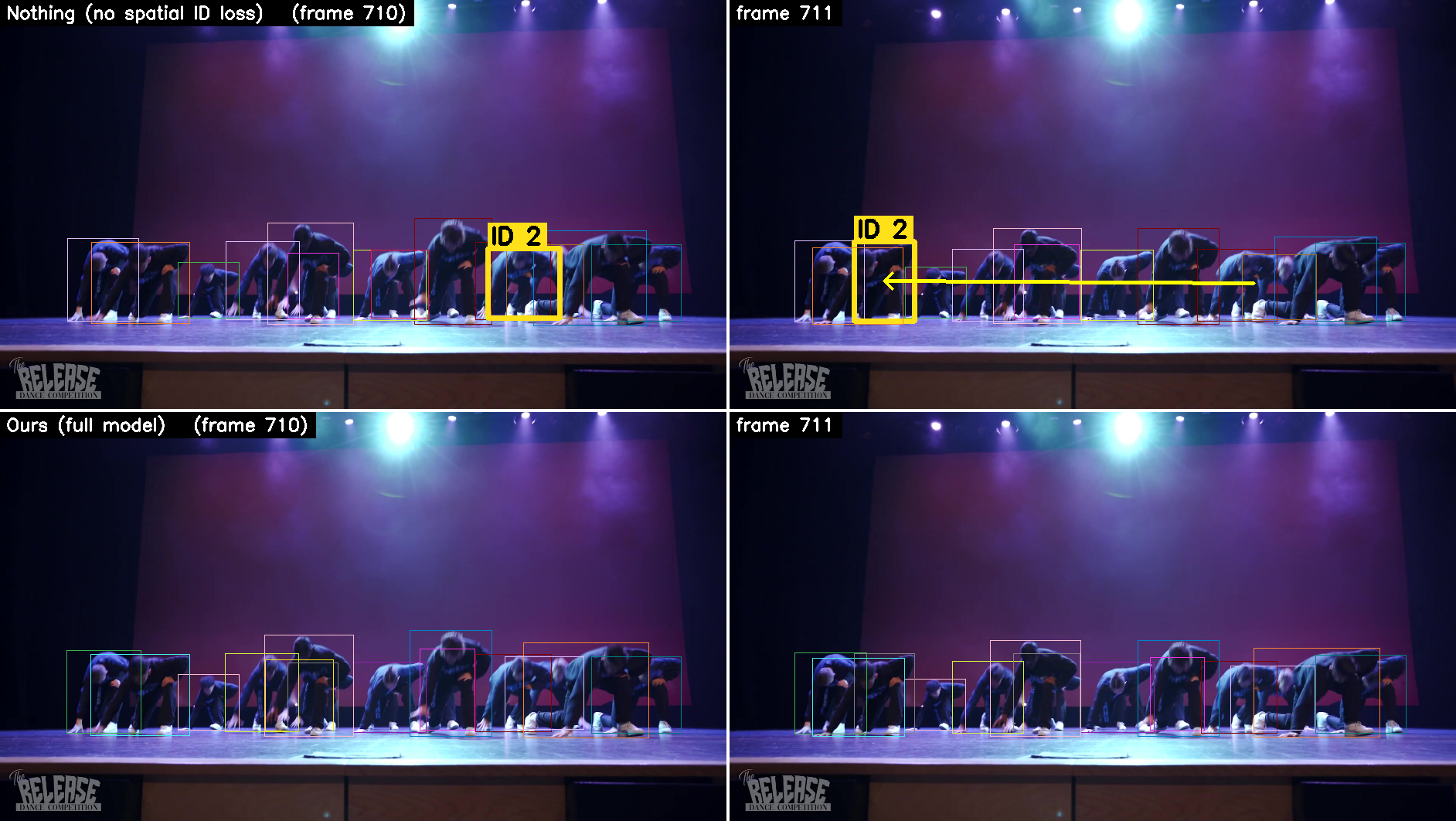}
  \caption{\textbf{A long-distance identity switch our spatial priors prevent.}
  DanceTrack \texttt{dancetrack0090}, consecutive frames $710$ and $711$.
  \emph{Top:} with the spatial priors disabled, identity \textbf{2} jumps
  $\approx\!975$\,px across the frame (yellow arrow), merging two distant dancers
  into a single track (\cref{sec:intro}). \emph{Bottom:} our full model keeps
  both identities stable.}
  \label{fig:longswitch}
\end{figure*}

\Cref{fig:longswitch} illustrates the long-distance switch raised in \cref{sec:intro}, the failure our spatial priors are designed to suppress. On a crowded DanceTrack sequence, the baseline trained without any of the spatial priors teleports an identity nearly across the full frame width between two consecutive frames, with no occlusion and no plausible motion path, merging two distant but near-identical dancers into a single track. Among the three priors, the Spatial ID Loss most directly targets this error by restoring the geometric cost that cross-entropy leaves out (\cref{sec:spatial_loss}). With all three priors active, our full model keeps both identities stable.

\section{Tracklet Confirmation Sweep}
\label[appendix]{sec:suppl_confirmation}

\Cref{tab:ablation_confirmation} reports the full tracklet-confirmation sweep summarized in \cref{sec:arch_ablations}. Requiring $N_{\text{init}}$ consecutive frames of consistent assignment before committing a tentative track leaves HOTA essentially unchanged, with only slight gains in IDF1 and MOTA, while steadily reducing identity switches, by $12.5\%$ on PersonPath22 ($5\,612 \!\to\! 4\,911$) and $28.3\%$ on DanceTrack ($3\,287 \!\to\! 2\,356$) from $N_{\text{init}}{=}1$ to $3$. Accuracy plateaus at $N_{\text{init}}{=}2$, and identity switches fall only marginally beyond it; we adopt $N_{\text{init}}{=}2$ as the smallest threshold that captures the gain.

\begin{table*}[t]
    \centering
    \small
    \setlength{\tabcolsep}{3pt}
    \begin{tabular}{@{}c@{\quad}cccccc@{\qquad}cccccc@{}}
        \toprule
        & \multicolumn{6}{c}{\textbf{PersonPath22}} & \multicolumn{6}{c}{\textbf{DanceTrack}} \\
        \cmidrule(lr){2-7} \cmidrule(lr){8-13}
        $N_{\text{init}}$
        & \textbf{HOTA} & \textbf{DetA} & \textbf{AssA} & \textbf{MOTA} & \textbf{IDF1} & \textbf{IDSW}$\downarrow$
        & \textbf{HOTA} & \textbf{DetA} & \textbf{AssA} & \textbf{MOTA} & \textbf{IDF1} & \textbf{IDSW}$\downarrow$ \\
        \midrule
        1 & 61.5 & 63.7 & 60.0 & 72.5 & 70.3 & 5\,612 & 64.3 & 73.7 & 56.4 & 81.9 & 67.7 & 3\,287 \\
        2 & \textbf{61.6} & \textbf{63.8} & \textbf{60.3} & 72.9 & \textbf{70.4} & 5\,069 & \textbf{64.5} & 74.0 & \textbf{56.5} & 82.9 & \textbf{68.8} & 2\,489 \\
        3 & \textbf{61.6} & \textbf{63.8} & \textbf{60.3} & \textbf{73.1} & \textbf{70.4} & \textbf{4\,911} & \textbf{64.5} & \textbf{74.2} & 56.3 & \textbf{83.2} & 68.3 & \textbf{2\,356} \\
        \bottomrule
    \end{tabular}
    \caption{\textbf{Effect of tracklet confirmation $N_{\text{init}}$.} MOTIP2-S on PersonPath22 test and DanceTrack val. $N_{\text{init}}$ is the number of consecutive frames a new tracklet must be matched before being confirmed. Best per metric and benchmark in \textbf{bold}.}
    \label{tab:ablation_confirmation}
\end{table*}

\section{Latency Comparison with Baselines}
\label[appendix]{sec:suppl_latency}

\cref{tab:latency_sota} compares per-frame latency against end-to-end baselines. MOTIP2 and MOTIP are profiled in PyTorch FP32 on a single V100; the FastTrackTr and PuTR figures (marked $^\dagger$) are estimated from the RTX 4090 measurements reported by~\cite{FastTrackTr}, scaled by MOTIP's V100/4090 latency ratio, so we read these cross-GPU comparisons as indicative rather than exact. At matched accuracy MOTIP2-S runs at $27.1$ ms against MOTIP's $86.9$ ms ($3.2\times$ faster) while reaching higher HOTA ($64.5$ vs $64.0$), and the lighter MOTIP2-N drops to $18.8$ ms.

\begin{table}[ht!]
    \centering
    \setlength{\tabcolsep}{3pt}
    \small
    \begin{tabular}{lc c c}
        \toprule
        \textbf{Method} & \textbf{HOTA} & \textbf{Latency (ms)} & \textbf{Resolution} \\
        \midrule
        \multicolumn{4}{l}{\textit{End-to-end}} \\
        FastTrackTr-R18-Dec3~\cite{FastTrackTr} & 47.4 & 17.1$^\dagger$ & 640$\times$640 \\
        FastTrackTr-Dec3~\cite{FastTrackTr}     & 51.2 & 27.2$^\dagger$ & 640$\times$640 \\
        PuTR~\cite{PuTR}                        & 54.5 & 39.9$^\dagger$ & 800$\times$1333 \\
        FastTrackTr (640)~\cite{FastTrackTr}    & 54.8 & 31.5$^\dagger$ & 640$\times$640 \\
        FastTrackTr (1333)~\cite{FastTrackTr}   & 56.9 & 53.6$^\dagger$ & 800$\times$1333 \\
        MOTIP~\cite{MOTIP}                       & 64.0 & 86.9            & 800$\times$1440 \\
        \midrule
        \multicolumn{4}{l}{\textit{End-to-end (Ours)}} \\
        \rowcolor{blueL!25} MOTIP2-N & 59.2 & 18.8 & 800$\times$1440 \\
        \rowcolor{blueL!25} MOTIP2-S & 64.5 & 27.1 & 800$\times$1440 \\
        \rowcolor{blueL!25} MOTIP2-M & 67.3 & 39.3 & 800$\times$1440 \\
        \rowcolor{blueL!25} MOTIP2-L & 68.0 & 54.8 & 800$\times$1440 \\
        \rowcolor{blueL!25} MOTIP2-X & 70.1 & 81.1 & 800$\times$1440 \\
        \bottomrule
    \end{tabular}
    \caption{\textbf{Latency comparison with end-to-end baselines.} HOTA on DanceTrack-val and per-frame latency (ms) in PyTorch on a single V100. $^\dagger$~Estimated from RTX 4090 measurements~\cite{FastTrackTr} scaled by MOTIP's V100/4090 latency ratio.}
    \label{tab:latency_sota}
\end{table}

\end{document}